\documentclass[journal]{IEEEtran}
\ifCLASSINFOpdf

\else

\fi
\usepackage{algorithm} 		
\usepackage{algorithmic} 	
\usepackage{bm}
\usepackage{booktabs}
\usepackage{graphicx}
\usepackage{amsmath}
\usepackage{cite}
\usepackage{amssymb}
\usepackage{multirow}

\usepackage{epstopdf}
\usepackage{exscale}
\usepackage{relsize}
\usepackage{color}
\usepackage{graphicx}
\usepackage{subfigure}
\usepackage[table]{xcolor}
\usepackage{amssymb,amsmath}
\usepackage{cases}
\usepackage{longtable}
\usepackage{rotating}
\usepackage{multirow}
 \usepackage{siunitx}
\usepackage{epstopdf}
\usepackage{cite}
\usepackage{float}
\usepackage[marginal]{footmisc}
\usepackage{stfloats}
\usepackage[colorlinks,
            linkcolor=black,
            anchorcolor=black,
            urlcolor=black,
            citecolor=black]{hyperref}
\usepackage[hyphenbreaks]{breakurl}

\input epsf
\begin{document}

\title{The Prospect of Enhancing Large-Scale Heterogeneous Federated Learning with Transformers}
\author{Yulan Gao\textsuperscript{$*$}, \IEEEmembership{~Member, IEEE,}
Zhaoxiang Hou\textsuperscript{$*$}, \IEEEmembership{}
Chengyi Yang, \IEEEmembership{}
Zengxiang Li, \IEEEmembership{}\\
and Han Yu, \IEEEmembership{~Senior Member, IEEE}
\thanks{Y. Gao and H Yu are with the School of Computer Science and Engineering, Nanyang Technological University (e-mail: yulan.gao@ntu.edu.sg, han.yu@ntu.edu.sg).}
\thanks{Z. Hou, C. Yang, and Z. Li are with Digital Research Institute, ENN Group, Beijing, China.(e-mail: lizengxiang@enn.cn).}
\thanks{\textsuperscript{*}The authors contributed equally.}}

\markboth{~}
{Shell \MakeLowercase{\textit{et al.}}: }
\maketitle
\begin{abstract}
Federated learning (FL) addresses data privacy concerns by enabling collaborative training of AI models across distributed data owners.
Wide adoption of FL faces the fundamental challenges of data heterogeneity and the large scale of data owners involved. In this paper, we investigate the prospect of Transformer-based FL models for achieving generalization and personalization in this setting. We conduct extensive comparative experiments involving FL with Transformers, ResNet, and personalized ResNet-based FL approaches under various scenarios. These experiments consider varying numbers of data owners to demonstrate Transformers' advantages over deep neural networks in large-scale heterogeneous FL tasks. In addition, we analyze the superior performance of Transformers by comparing the Centered Kernel Alignment (CKA) representation similarity across different layers and FL models to gain insight into the reasons behind their promising capabilities.
\end{abstract}
\begin{IEEEkeywords}
Transformer, federated learning, centered kernel alignment, large-scale.
\end{IEEEkeywords}
\IEEEpeerreviewmaketitle

\section{Introduction}
Federated learning (FL) as an innovative distributed learning for artificial intelligence that enables in-situ model training and testing by collaboratively computing over multiple heterogeneous devices ({\em a.k.a.} FL clients) \cite{yang2019federated,mcmahan2017communication}.
In contrast to traditional machine learning, FL trains a global model by aggregating and updating parameters computed locally by FL clients without sharing raw data, thereby reducing the risk of privacy leakage.
It has made impact on many fields such as smart cities \cite{zheng2022applications}, financial industries \cite{imteaj2022leveraging,byrd2020differentially,long2020federated}, and wireless communication \cite{gao2022multi,tran2019federated}.
Major industry players have launched their FL-based products.
For example, \texttt{Gboard}, Google's original virtual Keyboard app for iOS and Android launched in 2016, is adapting the FL process to improve the accuracy of word predictions while simultaneously maintaining data privacy of mobile devices.

Albeit the potential promising of FL in many applications, the deployment of FL in real-world data distributions remain poses several technical challenges, primarily due to the non-IID nature across heterogeneous devices.
Moreover, as the model size or the number of participants grows, the severity of non-IID characteristics becomes increasingly pronounced, which is a universal characteristic inherent in various industrial settings.
In this context, most Personalized FL (pFL) methods have been proposed to address two types of challenges: {\em  poor convergence on heterogeneous data} and {\em lack of solution personalization}.
For the first challenge, among the early contributions in this area, \cite{kairouz2021advances,mansour2020three} have employed the strategy of global model personalization to address the performance issue in heterogeneous data.
Conversely,  for the second challenge, the primary goal  is to train individual personalized FL models by modifying the model aggregation process.
These two types of PFL strategies have enhanced the accuracy, but they are predicated on specific assumptions.
Hence, the training process requires additional processing, resulting in high complexity (e.g., data augmentation, client selection, and regularization) and high model management costs \cite{tan2022towards}.
Consequently, PFL methods that involve additional training processing are not aligned with the requirements for deploying large-scale FL ecosystem.
Therefore, for large-scale FL systems, it is still imperative to conduct research to find solutions that are effective in handling non-IID issues and are yet cost-efficient.

Recently, with the strong applicable to heterogeneous data of Transformer architecture, there is an upsurge of revamping architectures in federated models rather than solely focusing on enhancing the optimization process of FL.
Due to the robustness provided by the combination of self-attention, large-scale pre-training, data augmentation, dropout, and layer normalization, transformers have become a popular choice for federated models where handling perturbations and distribution shifts is crucial \cite{vaswani2017attention,bahdanau2014neural,lyu2019advances}.
Specifically, \cite{hendrycks2020augmix,bhojanapalli2021understanding} have been proposed that the remarkable robustness of Vision Transformers (ViTs), which are a type of transformer architecture specifically designed for visual tasks, primarily stems from the self-attention mechanism of Transformers.
In the spirit of these works, a vast corpus literature has focused on
the integration of self-attention with Convolutional Neural Networks (CNN)-like architectures \cite{carion2020end,wang2019learning,wang2020axial}.
In a more recent study, the authors of \cite{qu2022rethinking} have conducted experiments on image classification \cite{krizhevsky2009learning,liu2015deep} and medical image classification \cite{kanungo2017detecting}, wherein they have replaced CNNs with Transformers.
The authors have inferred, based on empirical evidence, that ViTs demonstrate a high level of robustness across heterogeneous devices.

Despite empirical advances of ViTs in addressing non-IID data distributions, the issue of evaluating the performance of FLViT (FL with ViTs), especially in large-scale FL scenarios, remains unsolved.
So, concretely, we conduct numerous experiments of different neural architectures by varying the number of participants.
We further attempt to explore the underlying reasons of the superior performance of ViT in handling non-IID problems.
This is done by contrasting the Centered Kernel Alignment (CKA) representation similarity across varying layers and amongst a range of trained models.
To summarize, the key contributions and novelty of this paper can be outlined as follows:
\begin{itemize}
\item{ Extensive experiments reveal that the performance disparity between ViTFL and FL with ResNet grows with an increasing number of participants.
This further highlights the superior performance of ViTFL in dealing with non-IID problems compared to FL using ResNet.
It is noted that the FL with ResNet experiences a sharply decrease in global accuracy of $44.55\%$ as the number of participants increases from $10$ to $100$.
Additionally, Transformers can maintain high-performance baselines. }
\item{The comparison of global performance between ViTFL and two state-of-the-art pFL approaches, MOON and FedALA, applied to ResNet(50), reveals that ViT models exhibit remarkable capacity to handle non-IID problems. }
\item{We investigated the basis for ViTFL's superior performance by examining the representation similarity between different layers and models through CKA analysis.
Based on our experiments, we find that there is a strong uniform between both layers and devices in the early stages of training (i.e., when the epoch is small). }
\end{itemize}

In the remainder of this paper, the related work is given in Sec. II. The problem statement, personalized federated learning, and the CKA minibatch representational similarity are presented in Sec. III.
Sec. IV reports numerical experiments and corresponding discussion. Conclusions are presented in Sec. V.

\section{Related Work}
{\bf\em Federated Learning}.
As a distributed learning paradigm, FL solves the problem of data silos via multiparty collaboration on the premise of protecting data privacy.
With regard to enable distributed training across heterogeneous devices, a range of sophisticated approaches have been proposed to achieve the goals of improving FL model performance and training efficiency FedAvg \cite{mcmahan2017communication} and Cyclic Weight Transfer (CWT)\cite{chang2018distributed}.
Owing to the statistical heterogeneity of data incurred by the different data distributions and volume among participants, the biggest challenge for making FL a reality is to meet the soaring demand for data quality required to ensure model accuracy.
Despite the aforementioned challenge, some methods have been further proposed on the basis of FedAvg (e.g., FedProx \cite{li2020federated}, MOON \cite{li2021model}, FedALA \cite{zhang2022fedala}, etc.), demonstrating that it is feasible to improve the effect of Non-IID data.
More precisely, FedProx is a federated learning technique that introduces a proximal term locally to restrict local updates and bring them closer to the global model.
It has been observed to achieve more stable convergence and higher accuracy than FedAvg, especially under highly heterogeneous distributions.
Building on the principles of FedProx, the MOON approach incorporates model-level contrast loss into the local training objective to adjust for each participant's training.
The aim of MOON is to decrease the distance between the representation learned by the local model and the global model while increasing the distance between local model representations.
In non-IID FL, different participants require gaps in the global model information.
FedALA introduces an adaptive local aggregation method for FL, which enables accurate retrieval of the local information required by the local model from the global model.
Specifically, before each local training, FedALA aggregates the global model with the local model using an adaptive local aggregation module to automatically capture the necessary global model information.
Recent studies have shown that simply replacing the convolution-based architecture with the Transformer-based architecture can help reduce catastrophic model forgetting and better handle heterogeneous data  \cite{qu2022rethinking}.

{\bf\em Transformer}.
Transformers \cite{vaswani2017attention}  have emerged as a neural network architecture that has gained immense popularity in recent years, owing to their remarkable performance in a range of natural language processing (NLP) tasks.
The key principle behind the Transformer is self-attention, which allows the model to selectively focus on different parts of the input sequence to generate the output.
This is achieved through a mechanism that computes the importance of each input token based on its relationship with all other tokens in the sequence, rather than relying on fixed positional embeddings.
Now, the Transformer has emerged as a significant research avenue in the field of computer vision due to its exceptional performance and significant potential when compared to convolution-based architecture.
The authors of \cite{dosovitskiy2020image} have proposed a novel approach for image recognition by dividing the images into Patch via directly migrating the standard Transformer model to the image.
ViT, in comparison to the state-of-the-art convolution-based networks, achieves competitive results in image recognition tasks. However, its transfer learning process demands substantial computational resources and large-scale data to achieve optimal performance.
Subsequently, several variants have been proposed based on ViT to enhance its performance in various tasks.
For example, the authors of \cite{touvron2021training} have proposed data-efficient image transformers that address the issue of the extensive training data required for ViT by using a teacher-student model distillation approach to guide ViT.
The consistent resolution of the feature maps by ViT limits its capability to capture multi-scale information, thereby making it less optimal for tasks like object detection that require analysis of objects at different scales.
In order to address this limitation, a new approach called Swin-Transformer has been proposed in \cite{liu2021swin}.
Swin-Transformer leverages localized window self-attention and patch merging techniques to minimize computational overhead and establish a hierarchical structure, resulting in improved performance in various downstream computer vision tasks.
Moreover, it can serve as a versatile backbone architecture for computer vision models.

Extensive research has been conducted to investigate the internal representation structure of Transformer and CNN architectures, with a focus on understanding their relationship.
For example, Rao et al.  \cite{rao2022hornet} has found that Transformer models possess a capability to interact with higher-order spatial information compared to convolutional models.
Google has carried out further explorations, trying to confirm whether ViT and ResNet are consistent in the principle of processing images by the similarity of representations between models.
To be more precise, they have utilized a similarity metric called centered kernel alignment (CKA), as proposed in \cite{kornblith2019similarity}, to measure the degree of feature expression consistency between the networks.
The results suggest that ViTs and CNNs exhibit substantial dissimilarities in their respective feature representations and internal network structures \cite{raghu2021vision}.

\section{Methodology}
\subsection{Problem Statement}
We consider a typical scenario of FL which involves a FL server and a set of participants.
Let ${\cal N}:=\{1, 2, \ldots, N\}$ be the set of participants, where each participant $n~(\forall n\in{\cal N})$ is equipped with a local dataset ${\cal D}_n$, and $|{\cal D}_n|=D_n$.
Each data sample is represented as a pair of input and output, denoted as $\{{\boldsymbol x}_k, {\boldsymbol y}_k\}_{k=1}^{D_n}$.
And ${\boldsymbol x}_k$ corresponds to the label value of the input feature.
The local loss function for participant $n$, denoted as Eq. \eqref{eq:2}, is defined as the average loss over their individual training data.
And $L_k({\boldsymbol x}_k, {\boldsymbol y}_k, {\pmb\omega})$ describes the loss function.
\setcounter{equation}{0}
\begin{align}\label{eq:1}
{\cal L}_n({\pmb\omega})=\frac{1}{D_n}\sum\nolimits_{k=1}^{D_n}L_k({\boldsymbol x}_k, {\boldsymbol y}_k, {\pmb\omega}).
\end{align}
In addition, the optimization target of global federated learning process cloud be defined as:
\begin{equation}\label{eq:2}
\text{arg}\min_{{\pmb\omega}^{*}} {\cal L}({\boldsymbol w})=\sum\nolimits_{n=1}^{N} \frac{D_n}{D} * {\mathcal L}_n({\boldsymbol w}),
\end{equation}
where $D$ represents the combined size of all $N$ datasets, and  $D=\sum_{n\in{\cal N}}D_n$.
\subsection{Personalized Federated Learning}
pFL is an emerging field of study focused on developing techniques for training machine learning models in a decentralized manner, while prioritizing user privacy and customization.
This is achieved through the use of encryption techniques and data aggregation methods that allow the model to learn from the combined data of all users while keeping the individual data private.
There are several methods that have been proposed for pFL.
The FedBN \cite{li2020federated} and FedBABU \cite{oh2021fedbabu} methods do not aggregate all parameters,in particular, the Batch Normalization (BN) layer in FedBN and the classification head in FedBABU are excluded from the aggregation.
Furthermore, the pFedLA method employs only a hyper-network in its server without a global model.
To address these limitations, we have chosen two personalization methods, MOON and FedALA, which have demonstrated the ability to generate high-quality global models.
\setcounter{equation}{2}
\begin{figure*}[!t]
\begin{align}
\label{cka1} CKA_{\text{minibatch}}&=\frac{\frac{1}{k} \sum_{i=1}^{k} {HSIC}_{1}\left({\boldsymbol X}_{i} {\boldsymbol X}_{i}^{T}, {\boldsymbol Y}_{i} {\boldsymbol Y}_{i}^{T}\right)}{\sqrt{\frac{1}{k} \sum_{i=1}^{k} {HSIC}_{1}\left({\boldsymbol X}_{i} {\boldsymbol X}_{i}^{T}, {\boldsymbol X}_{i} {\boldsymbol X}_{i}^{T}\right)} \sqrt{\frac{1}{k} \sum_{i=1}^{k} {HSIC}_{1}\left({\boldsymbol Y}_{i} {\boldsymbol Y}_{i}^{T}, {\boldsymbol Y}_{i} {\boldsymbol Y}_{i}^{T}\right)}}. \\
\label{hsic} HSIC_{1}({\boldsymbol K}, {\boldsymbol L})&=\frac{1}{n(n-3)}\left(\text{Tr}(\tilde{{\boldsymbol K}} \tilde{{\boldsymbol L}})+\frac{\boldsymbol{1}^{T} \tilde{{\boldsymbol K}} {\boldsymbol 1} {\boldsymbol 1}^{T} \tilde{{\boldsymbol L}} {\boldsymbol 1}}{(n-1)(n-2)}-\frac{2}{n-2} {\boldsymbol 1}^{T} \tilde{{\boldsymbol K}} \tilde{{\boldsymbol L}} {\boldsymbol 1}\right).
 \end{align}
 \hrulefill
\end{figure*}

\subsection{CKA Minibatch Representational Similarity} \label{cka}
To visualize and understand the differences between Transformer and convolution-based architectural representations, we employ the CKA index $s({\boldsymbol X},{\boldsymbol Y})$, which can be used to compare representation similarities within and between neural networks. $s({\boldsymbol X},{\boldsymbol Y})$ ranges between 0 and 1, with 0 indicating no similarity at all and 1 indicating two identical.Unlike canonical correlation analysis (CCA) \cite{raghu2017svcca}, CKA can reliably identify the relationship of hidden layer representations between models trained with different initializations or networks of different widths.

Suppose there are two activation matrices, ${\boldsymbol X} \in \mathbb{R}^{m \times p_1} $ and ${\boldsymbol Y} \in \mathbb{R}^{m \times p_2}$ representing the representations of the hidden layers, where $m$ represents the number of samples and $p_1, p_2$ represent the number of neurons.
Inspired by \cite{nguyen2020wide}, the CKA index can be deduced from Eqs. \eqref{cka1} and \eqref{hsic}.
More precisely, this CKA index takes the activation matrix ${\boldsymbol X} \in \mathbb{R}^{m \times p_1} $ and ${\boldsymbol Y} \in \mathbb{R}^{m \times p_2} $ of two hidden layers as input and outputs the similarity of the matrices.
Moreover, $\tilde{{\boldsymbol K}}$  and  $\tilde{{\boldsymbol L}}$ are obtained by setting the diagonal terms of $\boldsymbol K$ and $\boldsymbol L$ to zero, $\text{Tr}(\cdot)$ indicates the trace of the matrix, and $\boldsymbol{1}$ represents an $n\text{-dimensional}$ column vector with element values of $1$.
We compute the average Hilbert-Schmidt independence criterion (HSIC) scores computed over $k$  minibatches to obtain the CKA index. Meanwhile, the unbiased estimator of HSIC is  used, so the value of CKA is independent of the batch size, as suggested in \cite{raghu2021vision,nguyen2020wide}.
\begin{table}[!htbp]
\caption{Generally Parameters of ViT and Resnet}
\centering
\label{tab:vit_resnet_parameters}
\scalebox{0.7}{
\begin{tabular}{l p{4cm} p{3.5cm}}
\hline\noalign{\smallskip}
 & ViT(s) & ResNet(50)  \\
\noalign{\smallskip}\hline\noalign{\smallskip}
Params	& 22.1 M	& 25.6 M \\
Image Size	& $224^2$	& $224^2$ \\
Batch Size	& 32	&32 \\
Client Epochs &	1	& 1 \\
Communication Rounds &	100 &	100 \\
Aggregation &	FedAvg	 &FedAvg \\
Pre-train &	ImageNet1k	 &ImageNet1k \\
\multirow{2}{*}{Optimizer}	& AdamW, lr=1e-5, wd= 0.05, momentum=0.9	& SGD, lr=0.03, wd= 0.0, momentum=0.9 \\
\noalign{\smallskip}\hline
\end{tabular}}
\end{table}
\begin{table}[!t]
\caption{The Accuracy of ViT and Resnet for Different Number of Participants Under Scenario 1 (S1) and Scenario 2 (S2).}
\centering
\label{tab:vit_resnet_results}
\scalebox{0.7}{
\begin{tabular}{c|c|>{\columncolor[RGB]{225,225,255}}c|>{\columncolor[RGB]{225,225,255}}c|c|>{\columncolor[RGB]{225,225,255}}c|>{\columncolor[RGB]{225,225,255}}c}
\hline
\multirow{2}{*}{Participants} & \multicolumn{3}{c|}{ViT(s)} & \multicolumn{3}{c}{ResNet(50)} \\
\cline{2-7}
~ & S1 & S2 (500) & S2 (1000) & S1 & S2 (500) & S2 (1000) \\
\hline
10	& 0.9672 & 0.8268 & 0.9001	& 0.9397 & 0.5372 & 0.7847 \\
20 	& 0.9568 & 0.8620 & 0.9228 	& 0.9178 & 0.4887 & 0.7909 \\
50	& 0.9422 & 0.8947 & 0.9422	& 0.7867 & 0.4936 & 0.7867\\
100	& 0.9064 & 0.9064 &0.9440	&\textcolor{red}{\bf 0.4942} & 0.4942 & 0.7818\\
\hline
\end{tabular}}
\end{table}
\section{Experiments and Discussion }
In this section, we conduct extensive experiments to evaluate ViTFL based on CIFAR-10.

\subsection{Experimental Setup}
{\bf\em CIFAR-10 Image Classification.}
CIFAR-10 \cite{krizhevsky2009learning} dataset comprises $60,000$ labeled images of small objects, divided into $10$ classes, with $50,000$ images for training and $10,000$ images for testing. Specifically, each class contains exactly $6,000$ images, making it a well-balanced dataset for classification tasks.
For our experiment, we created a non-IID data distribution by dividing the training set into $10, 20, 50$, and $100$ participants.
Moreover, each participant only contains four consecutive labels.
We investigate two experimental scenarios:
\begin{itemize}
\item{\bf Scenario 1:} the total volume of training data remains fixed.
The training set is equally distributed to multiple participants, so each participant receives a smaller portion of the training set as the number of participants increases.
That is to say, as the number of clients increases in a FL setting, it is likely that the degree of heterogeneity in the training dataset will also increase.
\item{\bf Scenarios 2:} the volume of data is consistent across all participants and is fixed.
For Scenario 2, we will conduct experiments where each participant is assigned a data allocation of either 500 or 1000.
Moreover, we utilize the $1000$ test sets as both local and global validation data.
\end{itemize}
In our work, we seek to evaluate and compare the performance of Transformer-based and convolution-based architectures in the above mentioned two scenarios.
To this end, we employed two models -- ViT(s) and Resnet(50) \cite{he2016deep} -- each designed with a comparable number of parameters.
These models are selected as representatives to enable us to draw conclusions about the two types of architectures.
In addition, several specific parameters of ViT and ResNet(50) are listed in Table \ref{tab:vit_resnet_parameters} (see \cite{qu2022rethinking} and references therein).

All FL experiments are conducted on a powerful server equipped with dual NVIDIA A100 GPUs, known for their cutting-edge performance in deep learning applications.
The simulations were carried out using the PyTorch 1.8 framework, which offers a versatile and scalable environment for training complex neural networks.
For the experiments involving $50$ and $100$ participants, we implement a model-sharing strategy among participants in order to reduce the server resource utilization.
\begin{figure*}
\centering
\includegraphics[width=2\columnwidth]{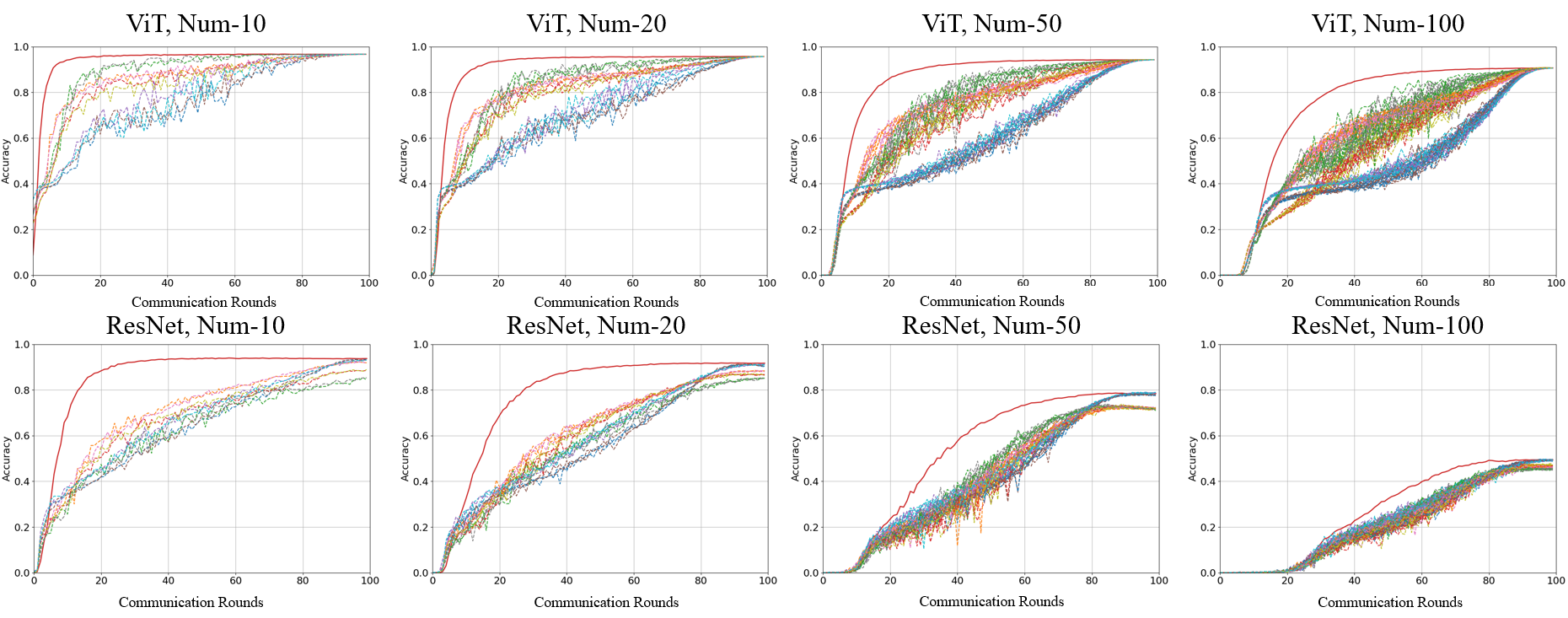}
\caption{Transformer (first row) and ResNet (second row)  test accuracy versus communication rounds on the CIFAR-10 dataset in Scenario 1. The solid red curve represents the accuracy of the server.}
\label{model_vit_resnet}
\end{figure*}
\begin{figure*}
\centering
\includegraphics[width=2\columnwidth]{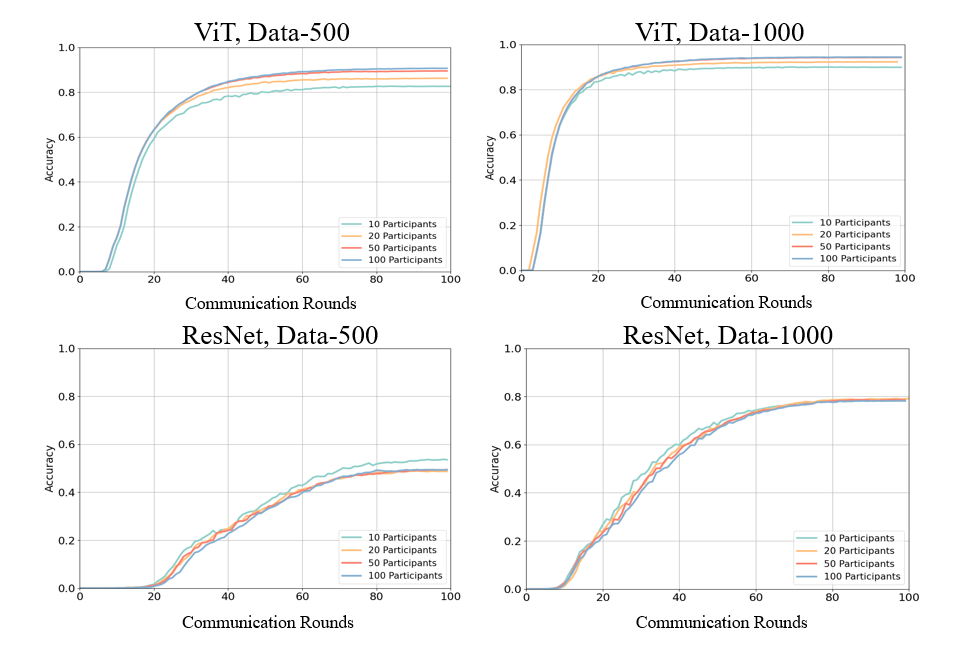}
\caption{Test accuracy  of Transformer (first row) and ResNet (second row) versus communication rounds with different number of participants $N\in\{10, 20, 50, 100\}$, for S2(500) and S2(1000).}
\label{model_vit_resnet_fix}
\end{figure*}

\begin{table*}[!t]
\caption{Summary of Personalized Federated Learning Methods}
\renewcommand\arraystretch{1.5}
\centering\label{tab:summary}
\scalebox{0.9}{
\begin{tabular}{|c|p{6cm}|p{6cm}|}
\hline
Method & Core Idea & Reasons for Not Conducting  A Comparison \\
\hline
\multirow{3}{*}{FedBN\cite{li2020federated}}
& Keep the local BN parameters not synchronized with the global model.
& The failure to update the parameters of BN layers can adversely affect the performance of the global model.\\
\hline
\multirow{2}{*}{Ditto\cite{li2021ditto} }& Incorporating personalized objectives into local models can strike a balance between personalized and global model performance.
& Ditto can result in both more robust and fairer models across a diverse set of attacks.\\
\hline
\multirow{2}{*}{MOON\cite{li2021model} }& Incorporate model-level contrast loss into the local training objective. & ~\\
\hline
\multirow{3}{*}{FedBABU \cite{oh2021fedbabu} }& Clients update the body of the model during federated training, and the head is fine-tuned for personalization during the evaluation process.
& The failure to update the parameters of the model's head can adversely influence the overall performance of the global model.\\
\hline
\multirow{3}{*}{pFedLA\cite{Ma_2022_CVPR} }& Use a dedicated hypernetwork per client on the server side, which is trained to identify the mutual contribution factors at layer granularity.
& At the server end, there exist only hypernetworks without a global model. \\
\hline
\multirow{2}{*}{FedALA\cite{zhang2022fedala} }
& Aggregate the global model with the local model using an adaptive local aggregation module. & ~\\ \hline
\end{tabular}}
\end{table*}
\begin{table}[!t]
\caption{The accuracy of ViT and Resnet with Personalized Methods for Different Number Participants in Scenario 1.}
\centering
\label{tab:resnet_pfl_results}       
\scalebox{0.8}{
\begin{tabular}{cccc}
\hline\noalign{\smallskip}
Participants & ViT & ResNet(50)+MOON & ResNet(50)+FedALA  \\
\noalign{\smallskip}\hline\noalign{\smallskip}
10	& {\bf 0.9672} & 0.9437    & 0.9567 \\
20 	& {\bf 0.9568} & 0.9046	& 0.9181 \\
50	& {\bf 0.9422} & 0.8301	& 0.7953 \\
100	& {\bf 0.9064} & \textcolor{red}{\bf 0.4242}	&\textcolor{red}{\bf 0.4968} \\
\noalign{\smallskip}\hline
\end{tabular}}
\end{table}
\subsection{Performance of Federated Learning with ViT(s)}
{\bf\em ViTFL Characteristics.}
We identify the performance characteristics of ViTFL with respect to the number of participants (i.e., the degree of heterogeneity in the training dataset) under Scenario 1 and Scenario 2.
To this end, Table \ref{tab:vit_resnet_results} represents the global accuracy results obtained by ViT and ResNet under two experimental scenarios, Scenario 1 and Scenario 2, while varying the number of participants from $10$ to $100$.
From the results, we observe that for both ViT and ResNet under Scenario 1, the global accuracy decreases, when the number of participants increases.
However, the reduction of the global accuracy for the ResNet network due to increasing the number of participants is more significant, than that for the ViT network.
More precisely, the ResNet network experiences a significant decrease in global accuracy of $44.55\%$ as the number of participants increases from $N=10$ to $N=100$.
By contrast, the ViT model demonstrates robustness as the number of participants increases from $N=10$ to $N=100$, with the global accuracy being maintained at over $90.00\%$.
Essentially, this phenomenon can be attributed to two factors.
Firstly, with an increase in the number of participants, there is a reduction in the amount of training data available for each participant, resulting in a decline in the quality of training.
Secondly, as the number of participants increases, the degree of heterogeneity in the training dataset also increases, further exacerbating the training quality.

To determine the primary factor contributing to the exceptional performance of ViT model, we conducted numerous experiments in Scenario 2.
For notational briefly, in the scenario where the volume of data contributed by each participant is $500$ ($1000$), it is denoted by S2(500) (S2(1000)).
The corresponding results are still presented in the purple part of Table \ref{tab:vit_resnet_results}.
It is clear from Table \ref{tab:vit_resnet_results} that the ViT model consistently outperforms the ResNet network in both S2(500) and S2(1000).
In S2(500), where each participant's dataset is relatively small, the global accuracy of ResNet exhibits a downward trend.
However,  the sensitivity of global accuracy to dataset heterogeneity decreases, as the number of participants increases.
Interestingly, for the ViT model in S2(500), the global accuracy increases with an increase in the level of dataset heterogeneity, as indicated by the rising values of $N$.
In addition, the behaviors of the ViT and ResNet in S2(1000) align with those observed in S2(500), respectively.
Combining the experimental results under Scenarios 1 and 2, we can infer that the robustness of ViT in dealing with dataset heterogeneity appears to be the primary factor in its superiority over the ResNet.
Based on our analysis, it reveals that the decrease in available training data to each participant is causing a decrease in performance for both ViT and ResNet in Scenario 1.
This in essence attributes to that when the available training data contributed by each participant reduces, the amount of diverse and complex patterns present in the data that the models can learn from also decreases, leading to lower performance.
\begin{figure*}
\centering
\includegraphics[width=2\columnwidth]{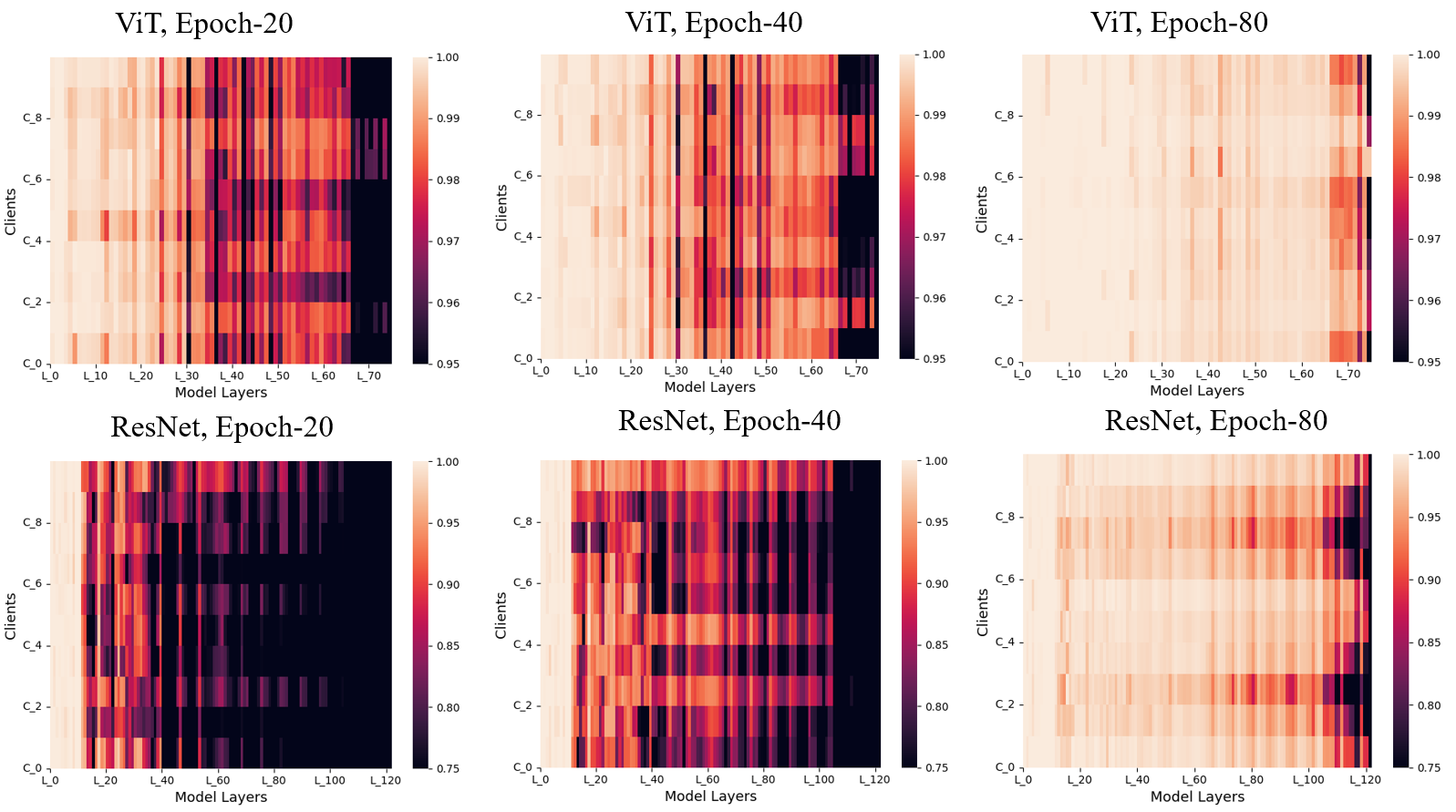}
\caption{Similarity of the same  layer between participants and server.The horizontal axis represents the remaining layers of ViT(s) and ResNet(50) network after removing some specific layers such as activation function layers and dropout layers.The vertical axis represents 10 participants.In the heatmap, the higher the CKA similarity index, the brighter the color.}
\label{layer_vit_resnet}
\end{figure*}
\begin{figure*}
\centering
\includegraphics[width=2\columnwidth]{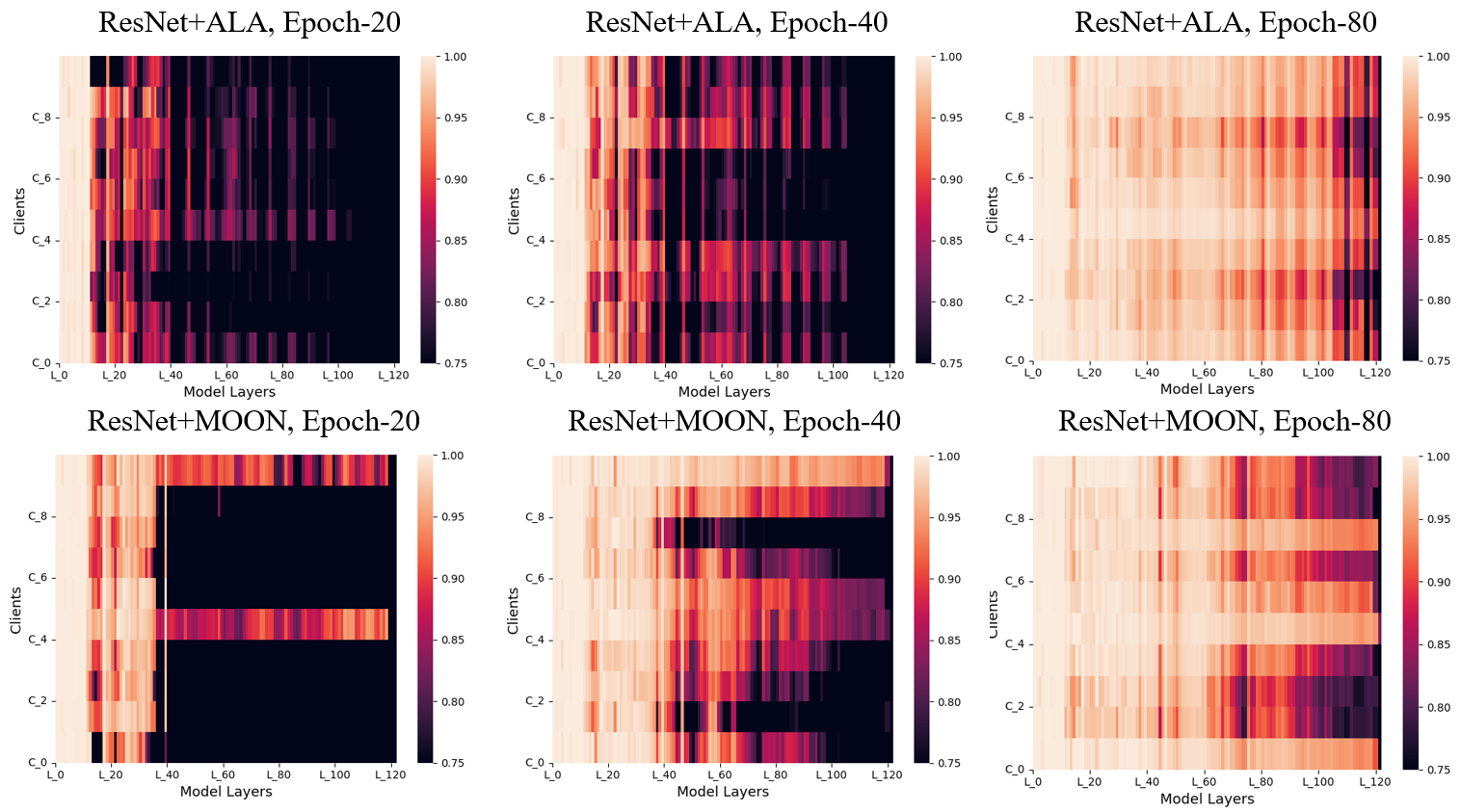}
\caption{Similarity of the ResNet same layer between participants and server with pFL methods.}
\label{layer_vit_resnet_pfl}
\end{figure*}

{\bf\em Convergence Speed.}
To further visualize the impact of ViT on convergence speed, we depict in Fig. \ref{model_vit_resnet} the global accuracy as a function of the communication rounds in Scenario 1.
In particular, for the sake of visualization and comprehensive analysis,  we report the behavior of the server and all participants for $N=10, 20, 50$, and $100$.
As can be seen, the convergence speed of both ViT and ResNet will slow down, as the number of participants $N$ increases, indicating that more communication rounds are required to achieve the same level of performance.
This can be attributed to the decrease in available training data per participant, which limits the models' ability to learn complex patterns and may result in less accurate or more generalizable representations.
In addition, for any given value of $N$, the ViT model convergence faster than ResNet.
From results we notice furthermore that, for the ViT, the convergence curves of the participants exhibit a clear clustering behavior, while for the ResNet such clustering behavior is not readily apparent.

In Fig. \ref{model_vit_resnet_fix}, we investigate the convergence speed of the ViT and ResNet in S2(500) and S2(1000) using different number of participants, measured by the test accuracy with respect to the communication rounds.
As seen in Fig. \ref{model_vit_resnet_fix}, for both ViT and ResNet, it is difficult to distinguish the convergence speed for $N=10, 20, 50, 100$ under a given scenario (S2(500) or S2(1000)).
We reasonably speculate that as the number of participants is increased, the resulting training dataset becomes larger and more heterogeneous.
And the positive and negative effects of increased dataset volume and dataset heterogeneity on model convergence performance may cancel each other out.
The results of the ResNet presented in Table \ref{tab:vit_resnet_results} for both S2(500) and S2(1000) provide strong evidence that supports our speculation.
\begin{figure*}[!t]
\centering
\includegraphics[width=2\columnwidth]{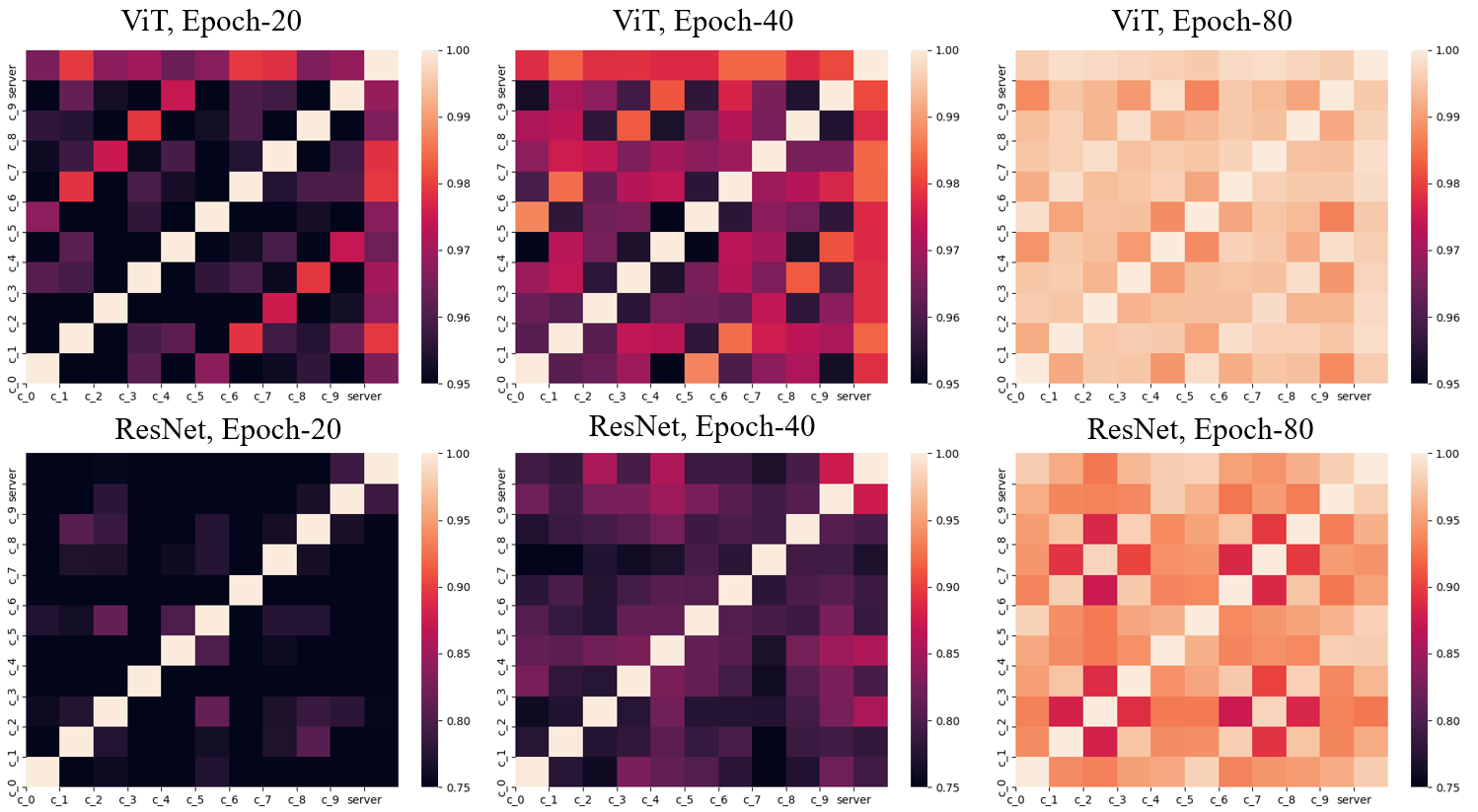}
\caption{The overall CKA similarity of each participant and server model.}
\label{cka_vit_resnet}
\end{figure*}
\begin{figure*}[!t]
\centering
\includegraphics[width=2\columnwidth]{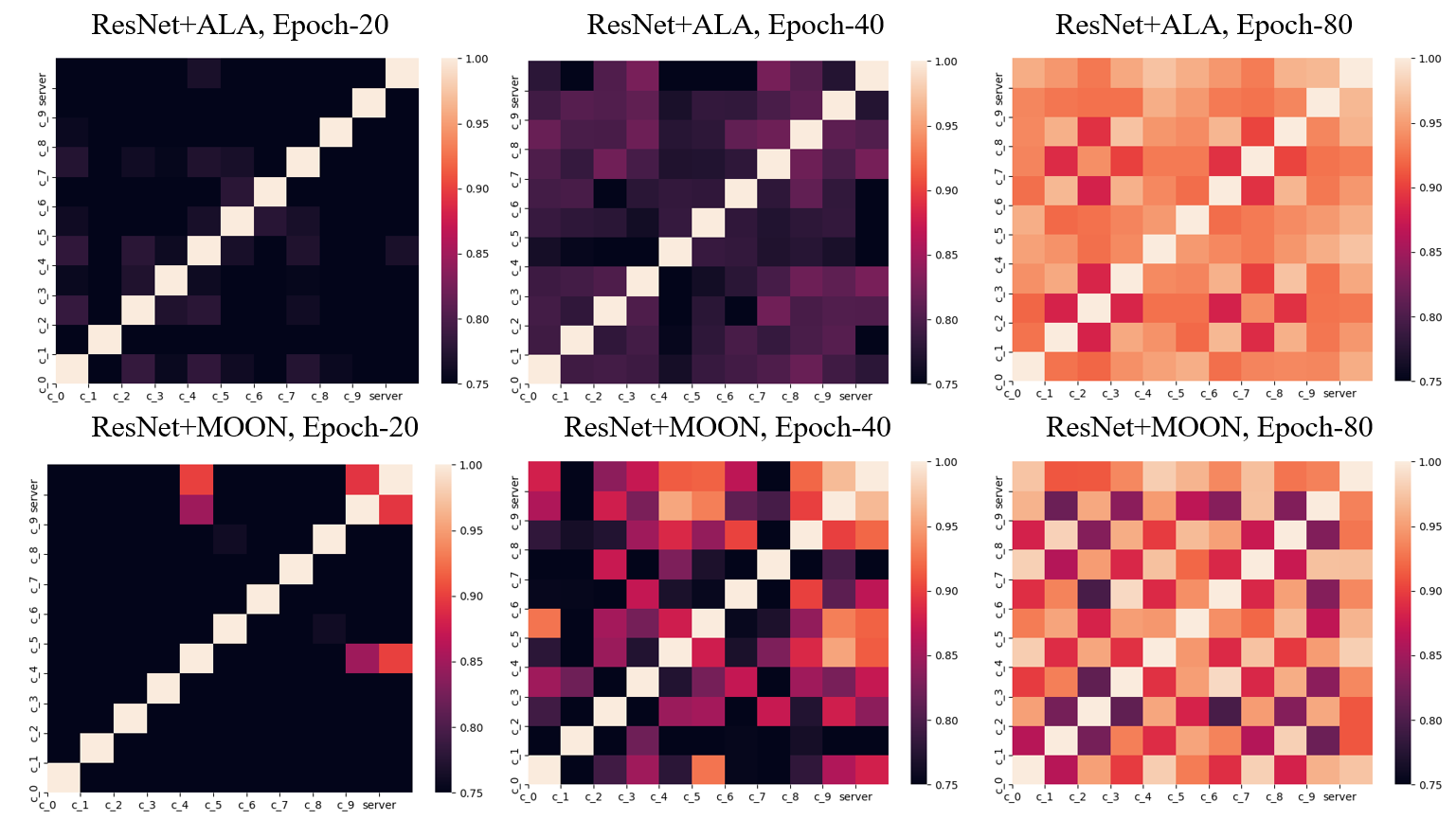}
\caption{The overall CKA similarity of Resnet each participant and server model  with pFL methods.}
\label{cka_vit_resnet_pfl}
\end{figure*}

\subsection{Benefits Verification of ViT(s)}
To gain insight into the the performance benefits of ViT(s), we conducted a comparative analysis by evaluating its test accuracy against two existing personalized federated learning methods.
In our comparative analysis, ResNet(50) is employed as the backbone network for the existing methods, while for our proposed approach, we utilized ViT(s).
This selection was made to ensure a fair comparison between the different methods while also highlighting the unique advantages of using ViT(s) in federated learning scenarios.
Specifically, we apply two personalized methods, MOON and FedALA,  to the ResNet(50) architecture, specifically designed to enhance the performance of models trained on non-IID  datasets.
In the following,  to simplify notation, the above mentioned comparison approaches are defined as ResNet(50)$+$MOON and ResNet(50)$+$FedALA\footnote{Here, Table \ref{tab:summary} provides a list of other typical pFL methods  along with the reasons for not using them as comparison algorithms.}, respectively.
The temperature coefficient of MOON is $0.5$ and the weighting factor $\mu$ for contrast loss is 5.
Likewise, set the sample random proportion(s) and the start of the adaptive layer of FedALA to be $100$ and $1$, respectively,  when its threshold is $0.05$.
As shown in  Table \ref{tab:resnet_pfl_results}, it is observed that the ViT enabled FL outperforms both ResNet(50)$+$MOON and ResNet(50)$+$FedALA, and the performance gap between them becomes larger with increasing the number of participants.
When the number of participants is small ($N\in\{10, 20\}$), the MOON and FedALA methods exhibit a more pronounced performance improvement effect.
Specifically, the FedALA method's performance improvement is particularly notable when there are $10$ participants involved.
The ViT achieves an accuracy that is only $1.05\%$ higher than that of the FedALA method.
However, as the number of participants increases to $100$, a sharp decline in performance becomes apparent.
The results presented in  Table \ref{tab:vit_resnet_results} for the ResNet, as well as Table \ref{tab:resnet_pfl_results} for the ResNet(50)$+$MOON, and ResNet(50)$+$FedALA, indicate that the current personalized methods (MOON and FedALA) have limitations in addressing heterogeneous datasets in  large-scale federated learning ecosystems.

\subsection{CKA Representation Similarity Analysis}
{\bf\em CKA Representation Similarity.}
To emphasize the significant influence of ViT models on the effectiveness of large-scale FL, Figs. \ref{layer_vit_resnet} and \ref{layer_vit_resnet_pfl} illustrate the comparisons of the performance of the CKA representation similarity between each participant and the FL server, on the same layer.
In particular, for the sake of conciseness, since all other participants show similar results, we only report the behavior of participants numbered $0$ to $9$, when the epochs are $20, 40$, and $80$, in the CIFAR-10 image classification task for the given total data volume.
Specifically, the corresponding CKA similarity scores can be deduced from: to compute HSIC, we employ a minibatch unbiased estimator, where each minibatch comprises $5$ images randomly selected from each category in the global validation dataset of CIFAR-10.
As can seen from these figures, the similarity between participants and the FL server for the ViT, ResNet, ResNet+ALA, and ResNet+MOON models tends to
increase as the training epoch progresses.
It is noted that the ViT model ensures that the similarity between participants and server models is maintained at a high level, i.e., $\text{CKA}\geq 95\%$, regardless of the specific layer being considered.
Besides observing the evolution of simalarity over training epochs, we also noticed significant differences in the heatmap structures of ViT, ResNet, ResNet+ALA, and ResNet+MOON models at the $20\text{~th}$ and $40\text{~th}$ epochs.
For the ResNet based models, the representations learned by participants and server from lower and higher layers are quite different.
By contrast, for the ViT model, both the participants and the server learned highly similar representations at lower layers (at epochs $20$ and $40$).
And the highly similar representations are learned across all layers, when the model converges.
The representations across all layers of ViT exhibit greater uniformity when compared to ResNet based models.

In Figs. \ref{cka_vit_resnet} and \ref{cka_vit_resnet_pfl}, we assessed and compared the  CKA similarity of each participant (numbered $0$ to $9$) and the server model.
For the small value of epoch, e.g., $20$ and $40$, all participants of ViT,  ResNet, ResNet+ALA, and ResNet+MOON models learned different representations.
When trained for a small number of epochs, such as $20$ and $40$, all the mentioned models produced distinct representations to each participant.
In addition, the CKA similarity of ViT and ResNet-based models cross all participants is high through as many epochs as possible (i.e., $80$).
In these figures, it can be observed that in addition to the subdiagonal of ViT's heatmap, there are also bright subdiagonals for the submatrices corresponding to epochs $20, 40$, and $80$.
These bright subdiagonal positions signify that the compared models have similar predictions for the dataset with the same labels during those epochs.


\section{Conclusion}
The broad adoption of FL is hindered by a decline in global accuracy due to non-IID data distribution.
To address this, we propose the incorporation of ViTs into FL frameworks, as a replacement for traditional CNNs.
Specifically, we showcase the exceptional ability of Transformers to handle non-IID problems while maintaining high-performance benchmarks.
To gain a comprehensive understanding of the benefits provided by Transformers, we conduct a series of extensive comparative experiments involving ViTFL, ResNet, ResNet+ALA, and ResNet+MOON models.
These experiments accommodate varying numbers of participants, facilitating a more robust assessment of Transformers' suitability in large-scale FL applications.
Additionally, we aim to understand the origin of this superior performance in large-scale scenarios by comparing the CKA representation similarity among different layers and diverse trained models.

Our extensive experimental results reveal significant patterns and intriguing phenomena that have broad implications for FL.
Given that Transformers are now commonly used for different types of data, we would, therefore, strongly encourage further investigations in this direction, particularly focusing on the novel challenges and opportunities that arise from studying FL within the context of Transformer-based network structures.



\end{document}